\documentclass[letterpaper, 10 pt, conference]{ieeeconf}  

\IEEEoverridecommandlockouts                              

\usepackage{graphics} 
\usepackage{graphicx}
\usepackage{epsfig} 
\usepackage{mathptmx} 
\usepackage{times} 
\usepackage{amsmath} 
\usepackage{amssymb}  
\usepackage[OT1, T1]{fontenc}
\usepackage{url}
\usepackage{booktabs}
\usepackage{cite}
\usepackage{multirow}
\usepackage{caption}
\usepackage[breaklinks=true,hidelinks]{hyperref}
\usepackage{cleveref}  

\usepackage{xcolor}
\definecolor{bluePoli}   {HTML}{0066B3}
\definecolor{structtitle}{HTML}{1B5E20}
\definecolor{structbord} {HTML}{C8E6C9}
\definecolor{structadd}  {HTML}{2E7D32}
\definecolor{lextitle}   {HTML}{0D47A1}
\definecolor{lexbord}    {HTML}{BBDEFB}
\definecolor{changedred} {HTML}{B71C1C}
\definecolor{xmlcol}     {HTML}{37474F}
\definecolor{labelcol}   {HTML}{607D8B}
\definecolor{bodycol}    {HTML}{212121}
\definecolor{steelblue}{HTML}{2E6DA4}
\definecolor{bordeaux} {HTML}{A63228}

\usepackage{helvet}   

\usepackage{tcolorbox}
\tcbuselibrary{skins, breakable}

\usepackage{listings}
\lstdefinestyle{yamlstyle}{
  language=,
  basicstyle=\fontsize{6.5}{8.5}\selectfont\sffamily\color{bodycol},
  keywordstyle=\bfseries\color{steelblue},
  stringstyle=\color{xmlcol},
  morestring=[b]",
  keywords={scene_analysis, target, destination,
             expanded_instruction, scene_context,
             expected_sequence},
  showstringspaces=false,
  breaklines=true,
  breakatwhitespace=true,
  columns=flexible,
  frame=none,
}

\lstdefinestyle{btxml}{
  language=XML,
  basicstyle=\fontsize{6.5}{8.5}\selectfont\sffamily\color{bodycol},
  keywordstyle=\bfseries\color{steelblue},
  stringstyle=\color{xmlcol},
  commentstyle=\color{labelcol}\itshape,
  tagstyle=\color{steelblue}\bfseries,
  showstringspaces=false,
  breaklines=true,
  columns=flexible,
  frame=none,
  morekeywords={root,BehaviorTree,Sequence,Fallback,Action,
                RetryUntilSuccessful,Condition,SubTree,Timeout},
}

\usepackage{tikz}
\usetikzlibrary{positioning, arrows.meta, calc}

\title{\LARGE \bf
Multimodal Behavior Tree Generation: A Small Vision-Language Model for Robot Task Planning
}

\author{
Riccardo Andrea Izzo$^{\dag*}$,
Cristiano Battistini$^{\dag*}$,
Gianluca Bardaro$^{\dag}$,
and Matteo Matteucci$^{\dag}$%
\thanks{$^{*}$Equal contribution}%
\thanks{$^{\dag}$All authors are with the Department of Electronics, Information, and Bioengineering,
Politecnico di Milano, Milan, Italy.
E-mail: {\texttt{cristiano.battistini@mail.polimi.it}},
{\texttt{\{riccardo.izzo, gianluca.bardaro, matteo.matteucci\}@polimi.it}}.}%
}

\begin{document}
\bstctlcite{IEEEexample:BSTcontrol}

\maketitle
\thispagestyle{empty}
\pagestyle{empty}

\begin{abstract}
Large language models have been widely used for robotic task planning, often taking advantage of representations such as Behavior Trees (BTs). Vision-Language Models (VLMs) have extended these works by grounding the generated plans in the observed scene. However, existing methods are either text-only or rely on large proprietary VLMs, while no dataset pairs visual observations and task instructions with executable and ROS2-compatible BTs. We address this gap with a multi-stage teacher pipeline that converts 1,622 Open X-Embodiment episodes into an augmented multimodal BT dataset containing 2,433 examples. We use this dataset to fine-tune compact and open-source VLMs, ranging from 500M to 4B parameters, using parameter-efficient fine-tuning (PEFT). We then evaluate the generated BTs offline in terms of syntactic correctness and by executing them on 15 household tasks in BEHAVIOR-1K. Our best model, Gemma-3 4B, achieves perfect BT validity and an 87\% success rate, outperforming Claude Opus 4.8 and approaching GPT-5, while running locally. Finally, our ablation studies show that adding visual observations increases task success from 40\% to 87\%, while data augmentation increases BT validity from 65\% to 100\%.
\end{abstract}

\section{INTRODUCTION}
\label{sec:introduction}
Robots are taking on increasingly complex roles in domestic and service
environments, going from logistics to personal care. Operating in unstructured
settings demands flexible task planning, capable of handling dynamic and
unpredictable surroundings where objects and human presence change
continuously. The development of embodied benchmarks such as
BEHAVIOR-1K~\cite{li2023behavior}, which defines over one thousand everyday
activities in a realistic simulation, confirms the breadth of this challenge.
Different formal representations exist to describe robotic
tasks~\cite{guo2023recent}, and among these, behavior trees (BTs)
have gained traction as a formalism that combines modularity with reactive
execution~\cite{colledanchise2018behavior,iovino2022survey}, being also widely adopted
in the ROS2 ecosystem with the BehaviorTree.CPP\footnote{\url{https://www.behaviortree.dev/}} library.
In this context, an increasingly explored direction is to automatically
synthesise these structured and directly executable representations from natural
language task instructions, reducing manual engineering effort and enabling
faster deployment across tasks and environments.

\begin{figure}[t]
  \centering
  \includegraphics[width=\columnwidth]{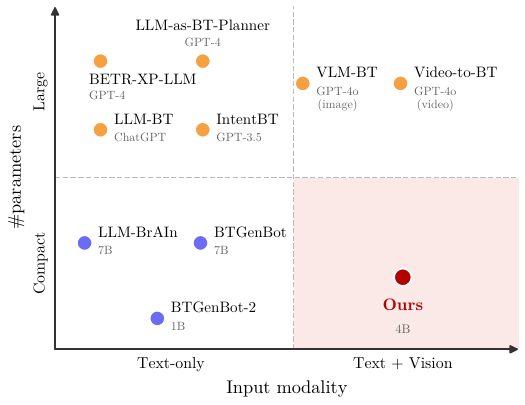}
  \caption{\textbf{BT generation methods by input modality and model scale.}
  Prior works either use text-only inputs or large closed-source models. The lower-right quadrant marks the gap addressed by this work.}
  \label{fig:bt_gen_positioning}
\end{figure}

Recently, Large Language Models (LLMs) have been applied to robotic planning,
as they can encode commonsense knowledge and interpret natural language
instructions~\cite{zhao2023large}. Along this line, a few works have shown
that LLMs can generate behavior trees in XML format from textual task
descriptions~\cite{lykov2024llm}, with one study demonstrating that compact fine-tuned models can produce valid behavior trees
for robotic tasks~\cite{izzo2024btgenbot}. However, text-only LLM approaches depend entirely on a textual description of
the task, with no direct perception of the scene. Indeed, a limitation
of BTGenBot~\cite{izzo2024btgenbot} was that the system had no access to visual
data: the robot could not observe its surroundings and instead had to act from
a text instruction alone, making it unable to adapt the plan to the actual state
of the environment. To address this lack of visual grounding, a few approaches have recently begun to integrate
Vision-Language Models (VLMs) for automatic BT generation~\cite{wake2025vlm,zhao2025video}.
Yet, as shown in Figure~\ref{fig:bt_gen_positioning}, these early attempts depend on large closed-source models and on curated prompt engineering. Furthermore, no existing dataset links visual observations and
natural language instructions to executable behavior trees. To the best of our
knowledge, no prior work has addressed behavior tree generation using compact and open-source VLMs.

In this work, we address these limitations by creating a multimodal BT dataset
that couples the RGB observations and task instructions with
executable BTs compatible with BehaviorTree.CPP. Starting from robotic episodes in Open X-Embodiment~\cite{o2024open}, a multi-stage teacher generation
pipeline produces BTs that are then used to train a VLM.

\begin{figure*}[t]
  \centering
  \includegraphics[width=\textwidth]{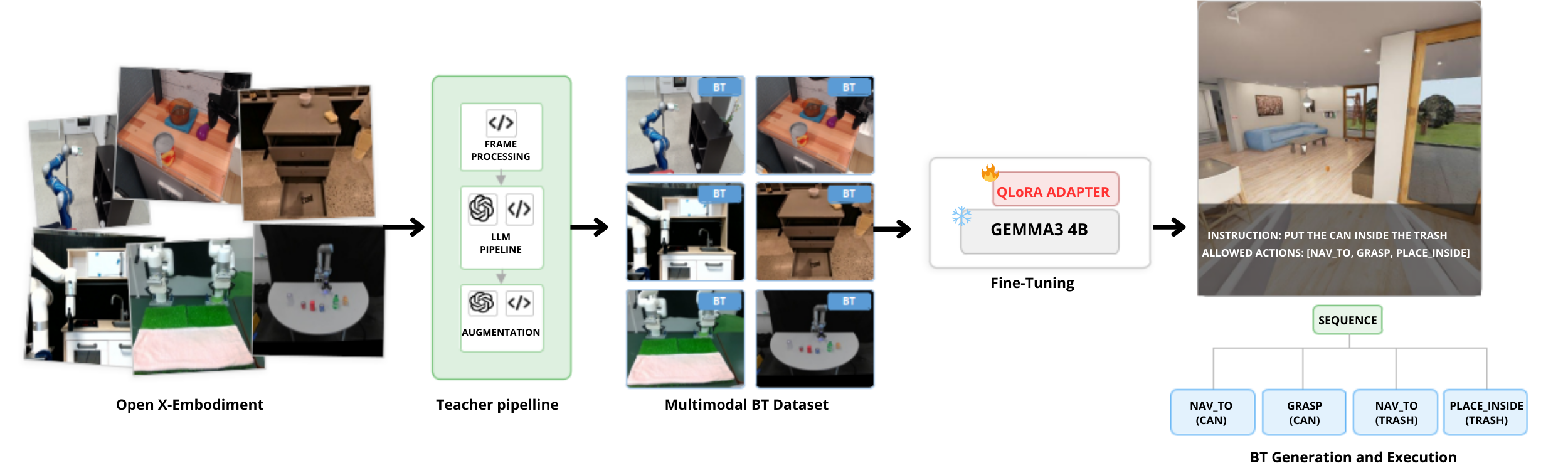}
  \caption{\textbf{Overview of our pipeline.} We derive image instruction pairs from Open X-Embodiment and use a multi-stage teacher pipeline to generate a multimodal BT dataset. Compact VLMs are then fine-tuned with QLoRA to output BTs compatible with BehaviorTree.CPP, using a single RGB image and a natural language instruction. Finally, generated BTs are executed in OmniGibson on the BEHAVIOR-1K household tasks to assess planning success.}
  \label{fig:hero}
\end{figure*}

The main contributions of this work are the following:
\begin{enumerate}
  \item \textbf{Multimodal BT Dataset}: an instruction-following dataset that pairs visual observations and natural language instructions with ROS2-compatible BTs.
  \item \textbf{VLM BT-Generator}: a compact vision-language model fine-tuned on our multimodal BT dataset via PEFT and suitable for deployment on ROS2 systems.
  \item \textbf{Empirical Analysis}: a systematic study showing how dataset augmentation, visual modality, model scale, and prompt guidance affect BT generation.
\end{enumerate}
We release model weights, code, and dataset\footnote{\url{https://github.com/AIRLab-POLIMI/multimodal-BT}}.

\section{Related Work}
\label{sec:related_work}

\subsection{Behavior Trees}
\label{subsec:behavior_trees}

A behavior tree (BT) is a plan representation based on a directed rooted
tree~\cite{colledanchise2018behavior}. Originating in the video
game industry, the idea was later adopted in robotics due to its modular
structure and its ability to react rapidly to changes in the
environment~\cite{iovino2022survey}. A BT has two types of nodes:
\emph{internal nodes}, which decide the execution order, and \emph{leaf
nodes}, which perform actions or check conditions. At a fixed frequency, a
signal called \emph{tick} starts at the root and travels downward; each
node that receives a tick runs and reports \textsc{Success},
\textsc{Failure}, or \textsc{Running}. The two main control-flow nodes
are \emph{Sequence}, which stops at the first child returning failure, and
\emph{Fallback}, which stops at the first child returning success. \emph{Decorator}
nodes such as \texttt{RetryUntilSuccessful} and \texttt{Timeout} wrap a
single child and modify its execution semantics. Leaf nodes are either
\emph{Actions}, which command the robot, or \emph{Conditions}, which
query the environment.

Prior studies comparing BTs to other representations, such as Finite State Machines, PDDL, and Decision Trees, highlight the advantages of BTs in terms of modularity, reactivity, and interpretability~\cite{biggar2021expressiveness}. 
These properties, combined with a unique XML representation, make them suitable for automatic generation and provide a structured output format that a language model can learn effectively.
Furthermore, BehaviorTree.CPP is the standard execution engine in ROS2~\cite{macenski2020marathon}, thus any model that produces valid XML in this format yields plans that can be loaded and run on a physical robot without intermediate conversion steps.

\subsection{Automatic BT Generation}
\label{subsec:bt_generation}

Existing approaches to automatic BT generation differ in the training methodology (fine-tuning on BT data versus prompting a general-purpose model) and in the input modalities (text-only or text paired with visual observations).

\noindent\textbf{Compact fine-tuned models.}
LLM-BrAIn~\cite{lykov2024llm} fine-tuned a 7B-parameter model on
8,500 synthetically generated task-description/tree pairs, reporting
no perceived difference between human-written and machine-generated
trees, though lacking syntactic or semantic validation.
BTGenBot~\cite{izzo2024btgenbot} addressed these limitations by training on
roughly 600 behavior trees from real open-source robotics
projects~\cite{ghzouli2023behavior}, each paired with a natural language
description produced by GPT-3.5. They then evaluated the generated output with
syntactic validation, simulation, and physical robot execution.
BTGenBot-2~\cite{izzo2026btgenbot} subsequently scaled down the model
to LLaMA-3.2 1B using QLoRA and increased the training set to 5,204
instruction/tree pairs through synthetic augmentation. This 1B model
surpassed its 7B predecessor in success rate at a lower inference cost.
All three systems are text-only and thus depend on a natural language scene
description as their primary source of state information for planning.

\noindent\textbf{Proprietary models.}
A parallel line of work using only text relies on prompt engineering with closed-source models. LLM-as-BT-Planner~\cite{ao2025llm} prompts GPT-4 with a structured scene
description and a formal action model to produce BTs for robotic assembly.
LLM-BT~\cite{zhou2024llm} expands the tree at execution time to recover
from unexpected disturbances, using a 3D camera and a recognition network
to build a textual semantic map that ChatGPT reads alongside the user
instruction. Although a perception module is present, the language model
itself still processes only text.
Two other works, IntentBT~\cite{chen2024integrating} and
BETR-XP-LLM~\cite{styrud2025automatic}, use the LLM to interpret the instruction,
while a separate classical planner constructs and expands the behavior tree.
As a result, the LLM does not directly output the BT structure.
The majority of these systems depend on closed-source APIs, and none is fine-tuned on a
specific BT dataset with visual input.

\begin{figure*}[t]
  \centering
  \includegraphics[width=0.93\textwidth]{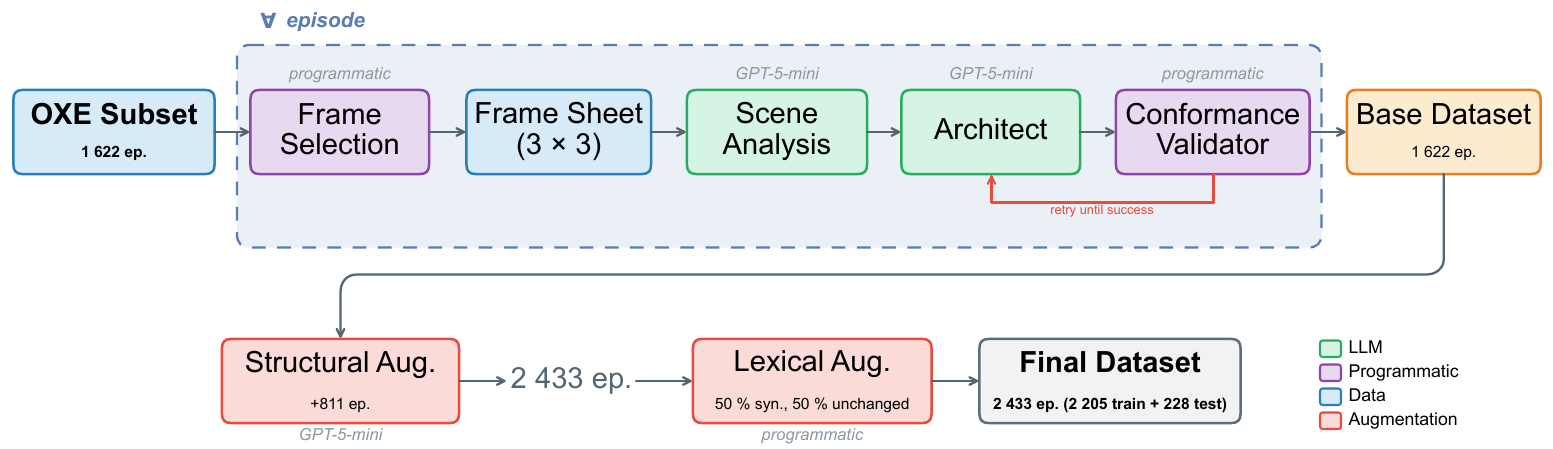}
    \caption{\textbf{Dataset generation pipeline.} Starting from an Open
    X-Embodiment subset (1,622 episodes), we build a $3{\times}3$ frame sheet as a temporally sparse episode summary and use GPT-5-mini to generate Scene Analysis
    and a linear BT XML plan, retrying the Architect step until it
    passes the Conformance Validator. This yields the base dataset (1,622 episodes).
    We then apply structural augmentation to 50\% of base episodes (+811), obtaining
    2,433 episodes, and apply lexical augmentation to the resulting set by
    replacing action names with the same probability.}
  \label{fig:teacher_pipeline}
\end{figure*}

\noindent\textbf{Adding vision.}
Vision-Language Models (VLMs) extend LLMs with a visual encoder, enabling
generation conditioned on both images and text.
VLM-BT~\cite{wake2025vlm} prompts GPT-4o with a textual instruction, a
map, and a skill list to generate a BT in JSON format. At execution time,
camera images are fed back to GPT-4o to evaluate visual condition nodes,
enabling reactive branching. Video-to-BT~\cite{zhao2025video} moves the visual input upstream: instead
of camera frames at execution time, the pipeline takes a human
demonstration video and uses GPT-4o, Gemini-1.5-flash, or Qwen-2.5VL-72B
to decompose it into reactive BT skeletons, while pretrained object
detection and pose estimation models handle perception~\cite{zhao2025video}.
Both systems, however, share two key limitations. First, they depend on
very large models that are not deployable on real robots, making them unsuitable for offline scenarios or resource-constrained hardware. Second, they do not operate zero-shot and rely heavily on prompt engineering.

As illustrated in Figure~\ref{fig:bt_gen_positioning}, no prior work occupies the intersection of compact
model size and visual input. This work fills that gap with a lightweight VLM that works zero-shot and provides ROS2 compatibility for integration with physical robotic platforms.

\section{Method}
\label{sec:method}

\subsection{Problem Formulation}
\label{subsec:problem_formulation}

Our goal is to generate an executable BT in XML format (compatible with BehaviorTree.CPP) that solves the instructed task given three inputs: (i) the current RGB observation,
(ii) a natural-language instruction, and (iii) a list of allowed
primitive actions, as shown in Figure~\ref{fig:hero}. During dataset construction, a large
\emph{teacher} model is used offline to generate training targets from an
episode-level visual summary, while a compact \emph{student} VLM is trained using only the single-frame observation available at test time.

The student model $f_\theta$ learns the mapping
\begin{equation}
\label{eq:bt_mapping}
  f_\theta \colon (\mathbf{I},\; t,\; \mathcal{A})
  \;\longrightarrow\;
  (\mathit{SA},\; \mathit{BT}),
\end{equation}
where $\mathbf{I} \in \mathbb{R}^{H \times W \times 3}$ is the RGB observation
(student input), $t$ is the instruction string, and $\mathcal{A}$ is the
task-specific list of allowed primitives provided in the user prompt.
The outputs are $\mathit{SA}$, a structured state analysis of the scene, and
$\mathit{BT}$, the generated BT in XML.

We define a fixed primitive library $\mathcal{P}$ of 22 actions, and require
$\mathcal{A} \subseteq \mathcal{P}$. The target XML is required to use only
actions from $\mathcal{A}$.

\subsection{Dataset Construction}
\label{subsec:dataset_construction}

\noindent\textbf{Source data.}
We leverage Open X-Embodiment~\cite{o2024open}, a large collection
of over one million robotic episodes across 22 embodiments. From this
collection, we select 23 datasets covering tabletop manipulation,
assistive control, mobile manipulation, and bimanual tasks. Only the visual
observations and the language instructions are retained, while all low-level
action sequences are explicitly discarded. This yields an Open X-Embodiment
subset of 1,622 episodes used for dataset generation (Figure~\ref{fig:teacher_pipeline}).

\noindent\textbf{Frame selection.}
Each episode is first sub-sampled using a temporal stride of 10 frames. We embed each
candidate frame with MobileNetV2~\cite{sandler2018mobilenetv2} and apply
K-center greedy~\cite{sener2017active} in the embedding space to select nine
visually distinct frames. The selected frames are sorted by timestamp and
arranged into a $3{\times}3$ frame sheet, resulting in a temporally sparse summary
of the episode.
This summary is used only offline by the teacher to infer task progression
when generating training targets $(\mathit{SA},\,\mathit{BT})$. For each episode, we store both the contact
sheet (teacher input) and the initial frame $\mathbf{I}$ (student input). The
student is trained on $\mathbf{I}$ only, matching the test-time operation where the
robot observes only the current scene.

\noindent\textbf{Teacher pipeline.} We split target generation into two calls to GPT-5-mini to produce
$(\mathit{SA},\,\mathit{BT})$ from the episode-level visual summary and the
instruction $t$. Both stages take as input the $3{\times}3$ frame sheet and the instruction $t$.

\emph{Scene Analysis} outputs $\mathit{SA}$ as a structured YAML block with five
fields: \texttt{target} (object to manipulate), \texttt{destination} (target
location), \texttt{expanded\_instruction} (instruction rewritten with grounded
object, location, and state details), \texttt{scene\_context} (initial scene
state), and \texttt{expected\_sequence} (natural language action plan). The
student is trained to output this YAML block, including
\texttt{expected\_sequence}, before the XML plan.

\emph{Architect} takes as input the frame sheet, the instruction $t$, and the
Scene Analysis output $\mathit{SA}$, and generates $\mathit{BT}$ as a linear
behavior tree XML. In the Architect prompt, GPT-5-mini is
given the full primitive library $\mathcal{P}$ (22 actions) and selects which
primitives to use for the current episode when synthesising the plan. Once the 
plan is generated, we extract from the XML the primitives that
actually appear in the tree and include them in the student prompt as the list of allowed actions $\mathcal{A}$ for that episode. Training with this
restricted list encourages the student to generate $\mathit{BT}$ using only
primitives in $\mathcal{A}$, and discourages actions that do not fit the episode.

A programmatic Conformance Validator checks that the XML is parseable by
BehaviorTree.CPP and that every \texttt{Action} belongs to $\mathcal{P}$. If
validation fails, the Architect stage is re-invoked until a conforming tree is
produced.

\noindent\textbf{Augmentation.}
Starting from the base dataset (1,622 episodes),
we apply augmentation in two steps.
\emph{Structural augmentation} is applied to 50\% of the base episodes (811/1,622).
For each selected episode, we use prompt templates to ask GPT-5-mini to output
both an augmented XML and an updated instruction that requests the same control-flow
construct. For example, if the instruction is augmented to request up to three
retries for a certain action, the XML is augmented by wrapping the corresponding
\texttt{Action} node with \texttt{RetryUntilSuccessful}. This yields 811 additional
episodes, resulting in 2,433 episodes before lexical augmentation. For each
structurally augmented sample, $\mathcal{A}$ is re-derived from the resulting XML.

\emph{Lexical augmentation} is then applied to the full set of 2,433 episodes.
With probability 0.5, we replace primitive identifiers with common synonyms
(e.g., \texttt{GRASP}$\rightarrow$\texttt{GRAB} or \texttt{PICK}). For placement
actions, we also generate variants that make the manipulated object explicit (in
addition to the destination); the object is inferred by tracking the most recent
\texttt{GRASP} in the tree. The remaining episodes are kept unchanged.
All lexical transformations update both the XML and the list of allowed actions
$\mathcal{A}$ in the user prompt, preserving input-output consistency and
teaching the model to condition action names on the list of allowed actions provided at inference time.

\begin{figure}[h]
\centering
{\sffamily   
\begin{tcolorbox}[
  enhanced, width=0.90\columnwidth,
  boxrule=0.5pt, colframe=bluePoli!40, colback=bluePoli!3,
  arc=2pt, outer arc=2pt,
  top=0pt, bottom=0pt, left=0pt, right=0pt,
]
  \begin{tcolorbox}[
    enhanced, colback=bluePoli!5, colframe=bluePoli!5,
    fonttitle=\centering\bfseries\scriptsize\color{black},
    title={\textcolor{bluePoli}{\textsc{Input} (User Turn)}},
    boxrule=0pt, arc=0pt,
    top=2pt, bottom=2pt, left=4pt, right=4pt,
    toptitle=1pt, bottomtitle=1pt,
  ]
  {\centering
    \includegraphics[width=0.55\linewidth]{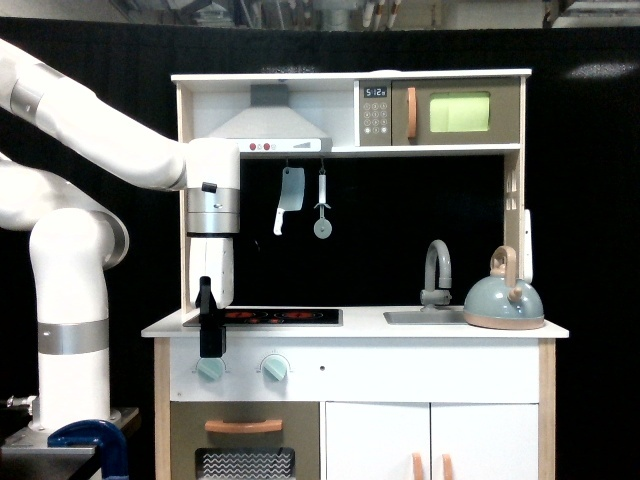}\par}
  \vspace{3pt}
  {\scriptsize\textbf{Instruction:} \textit{``Place the teapot on the stove''}}\\[1pt]
  {\scriptsize\textbf{Allowed actions:}
    \texttt{NAVIGATE\_TO}, \texttt{GRASP}, \texttt{PLACE\_ON\_TOP}}
  \end{tcolorbox}
  \vspace{-1pt}
  {\centering\scriptsize\bfseries
    \textcolor{bluePoli}{\textsc{Output} (Assistant Turn)}\par}
  \vspace{0pt}
  \begin{tcolorbox}[
    enhanced, colback=bluePoli!3, colframe=bluePoli!3,
    fonttitle=\bfseries\tiny\color{black}, title={State Analysis (YAML)},
    boxrule=0pt, arc=0pt,
    top=0pt, bottom=0pt, left=4pt, right=4pt,
    toptitle=0pt, bottomtitle=0pt,
    before skip=0pt, after skip=0pt,
  ]
  \begin{lstlisting}[style=yamlstyle, aboveskip=0pt, belowskip=0pt,
    breaklines=true, breakatwhitespace=true, columns=flexible]
scene_analysis:
  target: "teapot"
  destination: "stove"
  expanded_instruction: "Pick up the blue teapot and
    place it on the stove"
  scene_context: "Teapot on countertop right of sink;
    stove surface is clear."
  expected_sequence: "Approach teapot, grasp, move to
    stove, set it down."
  \end{lstlisting}
  \end{tcolorbox}
  \vspace{-1pt}
  \begin{tcolorbox}[
    enhanced, colback=bluePoli!3, colframe=bluePoli!3,
    fonttitle=\bfseries\tiny\color{black}, title={Behavior Tree (XML)},
    boxrule=0pt, arc=0pt,
    top=0pt, bottom=0pt, left=4pt, right=4pt,
    toptitle=0pt, bottomtitle=0pt,
    before skip=0pt, after skip=0pt,
  ]
  \begin{lstlisting}[style=btxml, aboveskip=0pt, belowskip=0pt,
    breaklines=true, columns=flexible]
<root main_tree_to_execute="MainTree">
  <BehaviorTree ID="MainTree">
    <Sequence>
      <Action ID="NAVIGATE_TO" obj="teapot"/>
      <Action ID="GRASP"       obj="teapot"/>
      <Action ID="NAVIGATE_TO" obj="stove"/>
      <Action ID="PLACE_ON_TOP" obj="stove"/>
    </Sequence>
  </BehaviorTree>
</root>
  \end{lstlisting}
  \end{tcolorbox}
\end{tcolorbox}
}  
\caption{\textbf{Training sample example.} The user turn provides the image
  and the task instruction; the assistant's turn generates the state analysis
  YAML followed by the XML BT.}
\label{fig:training_sample}
\end{figure}

\noindent\textbf{Final dataset.}
Each record is formatted as a two-turn conversation: the user turn provides the
initial frame $\mathbf{I}$, the task instruction $t$, and the list of allowed actions $\mathcal{A}$; the assistant turn contains $\mathit{SA}$ followed by
$\mathit{BT}$ (Figure~\ref{fig:training_sample}). At inference time, the fine-tuned
VLM produces the same two-part output, and evaluation metrics are computed on the XML
only. The final dataset contains 2,433 episodes (2,205 for training, 228 for testing):
1,622 base episodes with linear trees and 811 structurally augmented episodes.

\subsection{Fine-Tuning}
\label{subsec:finetuning}

We select and fine-tune three compact open-source VLMs: SmolVLM2-500M~\cite{marafioti2025smolvlm}, Qwen2.5-VL-3B~\cite{bai2025qwen2}, and Gemma-3 4B Vision~\cite{team2025gemma}. We adopt QLoRA~\cite{dettmers2023qlora}, which quantises the frozen
base weights to 4-bit NF4 precision and trains only low-rank
adapters~\cite{hu2022lora} in BFloat16, reducing memory consumption
by roughly four times compared to standard LoRA. LoRA adapters are injected into all linear layers
of each model, covering the language backbone, the visual encoder,
and the projection module. In our case, we used a rank $r=16$, scaling factor
$\alpha=16$, and no dropout. All three models are trained with an NVIDIA L4 GPU for
3 epochs with a learning rate of $2\times10^{-4}$ and a batch size of 16.

\section{Experimental Results}
\label{sec:experimental_results}

\subsection{Experimental Setup}
\label{subsec:experimental_setup}

We evaluate the successful execution of the generated plans inside OmniGibson~\cite{li2023behavior}, the
reference simulator of the BEHAVIOR-1K benchmark, built on NVIDIA
Isaac Sim. The simulated agent is R1, a bimanual mobile manipulator
with a holonomic base (3-DOF), a 4-DOF torso, two 6-DOF arms, and
parallel-jaw grippers. The RGB observation fed to the VLM is captured
from the left wrist-mounted camera (eye-in-hand perspective), ensuring that the
gripper remains consistently visible in the foreground of every frame.

All episodes employ symbolic execution with each primitive action that is
realised as an instantaneous state change rather than a physics-based
motor command. For instance, \texttt{NAVIGATE\_TO} teleports the robot next to the
target, \texttt{GRASP} attaches the object to the end-effector, and
placement primitives reposition the object in hand at the specified destination.
By eliminating low-level control variance and physics noise, any failed
episode can be attributed to errors in the plan, such as incorrect actions, wrong objects, or improper ordering, rather than to motor execution.

The task is considered solved if and only if every BDDL goal predicate
defined by the BEHAVIOR-1K activity is fully satisfied
(e.g.,\ \texttt{inside(drill,\,toolbox)} or
\texttt{$\lnot$open(toolbox)}). Crucially, no partial credit is awarded. 
Inference runs on an NVIDIA RTX 5080 while simulation runs on a separate machine with an NVIDIA RTX 3080Ti.

Our three fine-tuned models are compared against their base counterparts
as well as top-performing proprietary models across two evaluation stages: offline metrics assess structural correctness and instruction grounding;
simulation execution on 15 BEHAVIOR-1K tasks determines whether the generated plans succeed on real household activities.

\subsection{Offline BT Synthesis}
\label{subsec:offline_eval}

Models are evaluated on a test split comprising 228 samples
(152~linear, 76~with decorators).

\begin{table}[t]
  \centering
  \caption{Structural metrics on the held-out test split ($N=228$).}
  \label{tab:offline_results}
  \scriptsize
  \begin{tabular}{lccc}
    \toprule
      & \textbf{SmolVLM2} & \textbf{Gemma-3} & \textbf{Qwen2.5-VL} \\
    \midrule
    \multicolumn{4}{l}{\textit{Adapter (fine-tuned)}} \\
    \midrule
    Inference time (s)    & $12.7{\pm}24.4$ & $20.4{\pm}5.5$     & $17.2{\pm}4.9$   \\
    XML validity (\%)    & 88.60           & 100.00             & 100.00           \\
    BT-CPP validity (\%) & 87.72           & 100.00             & 100.00           \\
    Structural compliance (\%) & 66.67      & 96.93              & 94.74            \\
    Linear                & 142/152         & 152/152            & 146/152          \\
    Decorator             & 10/76           & 69/76              & 70/76            \\
    Action Jaccard        & $0.886{\pm}0.197$ & $0.971{\pm}0.129$ & $0.984{\pm}0.079$ \\
    \midrule
    \multicolumn{4}{l}{\textit{Baseline (no fine-tuning)}} \\
    \midrule
    Inference time (s)    & $39.0{\pm}31.6$ & $104.6{\pm}114.2$  & $13.9{\pm}15.1$  \\
    XML validity (\%)    & 27.19           & 17.54              & 82.89            \\
    BT-CPP validity (\%) & 0.00            & 0.00               & 0.00             \\
    \bottomrule
  \end{tabular}
\end{table}

\emph{Structural compliance} evaluates whether the structure
of the generated tree matches the ground truth. For each pair
$(o_i, g_i)$ of generated and reference BTs, we extract the unique set of decorator tags present from the vocabulary $\mathcal{D} =
\{$\texttt{RetryUntilSuccessful}, \texttt{Fallback},
\texttt{Condition}, \texttt{Timeout}, \texttt{SubTree}$\}$ and
require exact set equality:
\begin{equation}
\label{eq:structural_match}
  \textsc{StructMatch}(o_i,\,g_i) =
  \begin{cases}
    1 & \text{if } \mathcal{D}(o_i) = \mathcal{D}(g_i),\\
    0 & \text{otherwise.}
  \end{cases}
\end{equation}
This metric quantifies the model's ability to adhere to user constraints: a plain sequence when no special construct is required,
or the appropriate decorator when the prompt contains a specific cue
(e.g.,\ ``retry if the grasp fails'' $\to$ \texttt{RetryUntilSuccessful}).

\emph{Action Jaccard} similarity measures the overlap between the sets
of action primitives present in the generated and reference trees:
\begin{equation}
\label{eq:action_jaccard}
  J(o_i,\,g_i) =
  \frac{|\,\mathcal{A}(o_i)\;\cap\;\mathcal{A}(g_i)\,|}
       {|\,\mathcal{A}(o_i)\;\cup\;\mathcal{A}(g_i)\,|},
\end{equation}
where $\mathcal{A}(\cdot)$ denotes the set of distinct action
primitives extracted from a tree, explicitly excluding structural tags and decorator nodes.  The index ranges from 0 (i.e., no action in common) to 1
(i.e., identical action vocabularies). 

As shown in Table~\ref{tab:offline_results}, without fine-tuning, none of the base models
delivers a single valid BT. The addition of a LoRA adapter closes this gap
entirely for Gemma-3 and Qwen2.5-VL, while SmolVLM2 retains a marginal
failure rate. Fine-tuning also
significantly cuts inference time. While base models generate long and unstructured text before reaching a stop token, the fine-tuned models produce well-formatted XML output.
Table~\ref{tab:offline_results} also reveals a clear insight: all three fine-tuned models handle linear trees well, but only the two larger models reliably reproduce control-flow constructs. Specifically, Gemma-3 achieves the highest structural score, whereas Qwen2.5-VL excels on action precision. 

\subsection{Simulation Benchmark and Main Results}
\label{subsec:simulation_eval}

While offline metrics provide structural validation and syntactic correctness, they cannot verify whether a generated plan achieves its goal in the embodied world. 
Consequently, generated plans may remain structurally close to the reference while being
non-executable due to violated preconditions (e.g.,~\texttt{PLACE\_INSIDE} before \texttt{OPEN}). 
To validate this, we execute the generated BTs inside the OmniGibson simulator on 15 BEHAVIOR-1K tasks,
measuring success rate via BDDL goal satisfaction, and compare the
three fine-tuned models, their base counterparts, with off-the-shelf models such as GPT-5 and Claude Opus 4.8.

\noindent\textbf{Tasks and prompting.}
A suite of fifteen BEHAVIOR-1K activities was selected to
cover a progressive range of planning complexity, as described in Table \ref{tab:task_catalog}.
The suite spans seven \emph{Easy} tasks requiring linear sequences, five \emph{Medium} tasks involving container states, and three \emph{Hard} tasks requiring nested containment or implicit physical preconditions.
Each BEHAVIOR-1K activity is associated with a set of BDDL goal predicates that formalise the desired final state (e.g.,\ \texttt{inside(drill,\,toolbox)}, \texttt{$\lnot$open(fridge)}).

\begin{table}[t]
  \centering
  \caption{BEHAVIOR-1K set of selected tasks.}
  \label{tab:task_catalog}
  \scriptsize
  \begin{tabular}{lll}
    \toprule
    \textbf{Task} & \textbf{Difficulty} & \textbf{Type} \\
    \midrule
    turning on radio            & Easy   & State-change \\
    picking up trash            & Easy   & Repeated pick-and-place \\
    setting mousetraps          & Easy   & Multi-object placement \\
    hiding Easter eggs          & Easy   & Shared-target placement \\
    bringing in wood            & Easy   & Repeated transport \\
    moving boxes to storage     & Easy   & Stacking \\
    tidying bedroom             & Easy   & Chained spatial \\
    \midrule
    picking up toys             & Medium & Many objects, one container \\
    putting up Christmas dec.   & Medium & Many objects, many surfaces \\
    preparing lunch box         & Medium & Container, mixed sources \\
    bringing water              & Medium & Container-mediated \\
    outfit a basic toolbox      & Medium & Repetitive container \\
    \midrule
    putting away Halloween dec. & Hard   & Mixed placement, open/close \\
    loading the car             & Hard   & Nested containment \\
    carrying in groceries       & Hard   & Strict ordering \\
    \bottomrule
  \end{tabular}
\end{table}

We test each model under two prompting strategies.
\emph{Chain-of-Thought} (CoT) includes a numbered natural language
\texttt{Workflow} that decomposes the task into the exact sequence of
primitives (no XML examples are provided).
\emph{Zero-Shot} (ZS) states the goal as a natural language instruction
with no explicit decomposition; the model must infer the full action
sequence from the instruction and the scene image alone. Three metrics
are reported: \emph{BT Valid} (XML syntactic validity and compatibility with BehaviorTree.CPP), \emph{SR} (success rate, first attempt satisfies all BDDL predicates),
and \emph{Pass@3} (at least one success in three attempts).

\noindent\textbf{Results.}
In Table~\ref{tab:aggregate_results}, fine-tuning appears to be essential as, without adaptation, no base model produces a single valid tree. After fine-tuning, Gemma-3
and Qwen2.5-VL achieve 100\% BT validity alongside strong planning performance,
with Gemma-3 obtaining 87\% SR and 93\% Pass@3 under
CoT prompting.
The higher Pass@3 relative to SR reflects stochastic variation in VLM generation, with each attempt that regenerates the BT and executes it after resetting the simulator.
However, the benefits of fine-tuning do not scale linearly with model capacity, revealing a qualitative threshold between 500M and 3B parameters.
SmolVLM2 achieves acceptable format validity but fails catastrophically to generalise to the BEHAVIOR-1K domain, where its
BT validity drops to 7\% under CoT. Once a model crosses
this parameter threshold, failures shift from syntactic to semantic, producing structurally valid XML that encodes an incorrect plan. 

\begin{figure}[t]
  \centering
  \includegraphics[width=0.95\columnwidth]{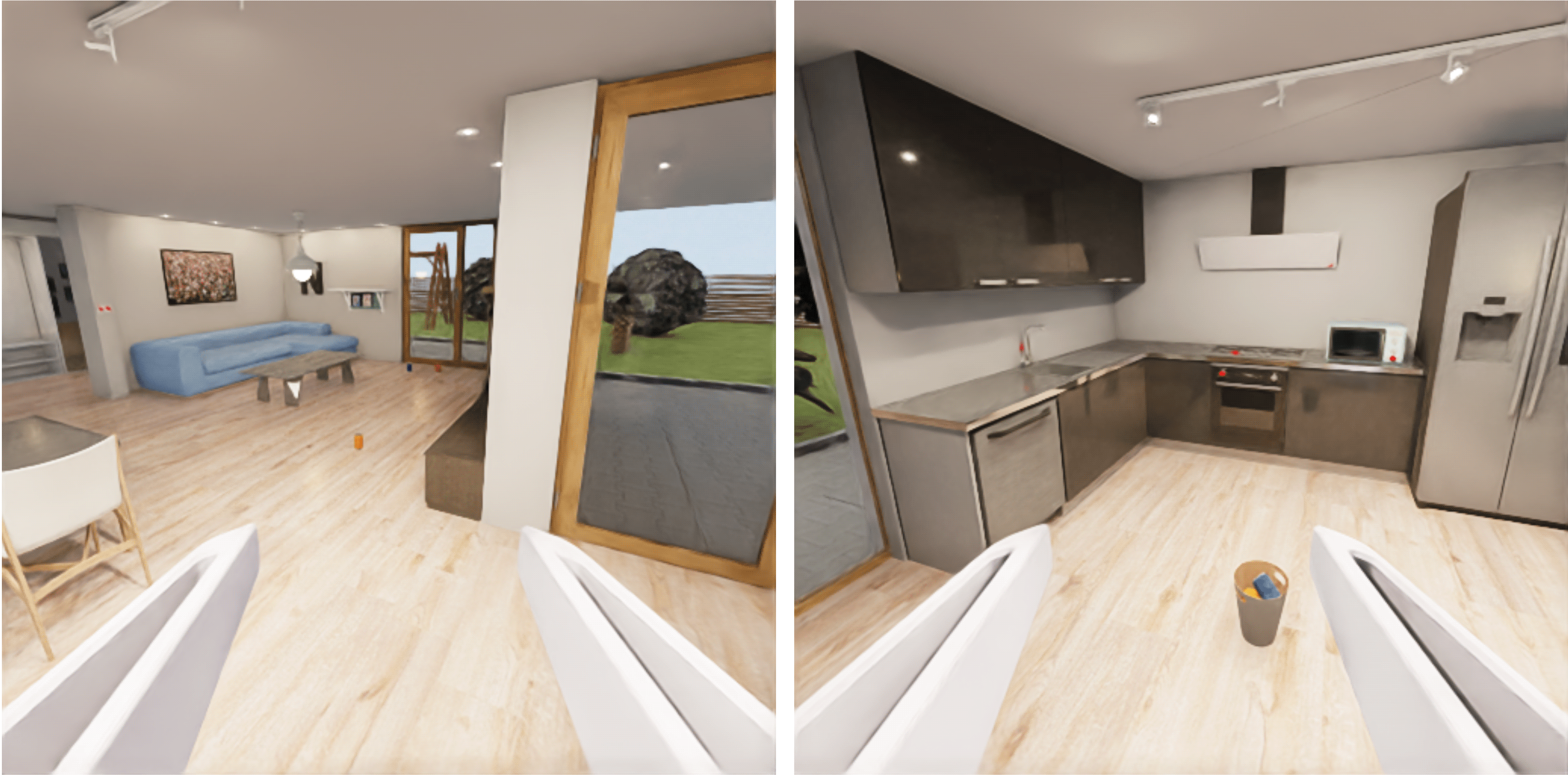}
  \vspace{1pt}

    {\sffamily
    \begin{tcolorbox}[enhanced, width=\columnwidth,
      boxrule=0.5pt, colframe=bluePoli!40, colback=bluePoli!3,
      arc=2pt, outer arc=2pt,
      top=0pt, bottom=0pt, left=0pt, right=0pt,
      before skip=0pt, after skip=6pt]
    
      \vspace{2pt}
      {\centering\scriptsize\bfseries
        \textbf{Gemma-3 4B}~{\normalsize\textbf{\textcolor{green!60!black}{\checkmark}}}~
        {\normalfont\scriptsize\textit{picking up trash}}
        \par}
      \vspace{1pt}
    
      \begin{tcolorbox}[
        enhanced, colback=bluePoli!3, colframe=bluePoli!3,
        fonttitle=\sffamily\bfseries\tiny\color{black},
        title={State Analysis (YAML)},
        boxrule=0pt, arc=0pt,
        top=0pt, bottom=0pt, left=4pt, right=4pt,
        toptitle=0pt, bottomtitle=0pt,
        before skip=0pt, after skip=0pt,
      ]
      \begin{lstlisting}[style=yamlstyle,
        basicstyle=\fontsize{5.5}{6.5}\ttfamily,
        backgroundcolor={}, % IMPORTANT: let tcolorbox provide the background
        aboveskip=0pt, belowskip=0pt,
        breaklines=true, breakatwhitespace=true, columns=flexible]
    scene_analysis:
      target: "can_of_soda"
      destination: "trash_can"
      expanded_instruction: "Pick up the three soda cans and place them in the kitchen trash can."
      scene_context: "Trash can open; three soda cans on living-room floor; not held."
      expected_sequence: "For each can: navigate, grasp, navigate to trash can, place inside."
      \end{lstlisting}
      \end{tcolorbox}
    
      \vspace{-1pt}
    
      \begin{tcolorbox}[
        enhanced, colback=bluePoli!3, colframe=bluePoli!3,
        fonttitle=\sffamily\bfseries\tiny\color{black},
        title={Behavior Tree (XML)},
        boxrule=0pt, arc=0pt,
        top=0pt, bottom=0pt, left=4pt, right=4pt,
        toptitle=0pt, bottomtitle=0pt,
        before skip=0pt, after skip=0pt,
      ]
      \begin{lstlisting}[style=btxml,
        basicstyle=\fontsize{5.5}{6.5}\ttfamily,
        backgroundcolor={}, % IMPORTANT: let tcolorbox provide the background
        aboveskip=0pt, belowskip=0pt,
        breaklines=true, columns=flexible]
    <Sequence>
      <Action ID="NAVIGATE_TO"  obj="can_of_soda_1"/>
      <Action ID="GRASP"        obj="can_of_soda_1"/>
      <Action ID="NAVIGATE_TO"  obj="trash_can"/>
      <Action ID="PLACE_INSIDE" obj="trash_can"/>
      <!-- repeat for can_of_soda_2 and can_of_soda_3 -->
    </Sequence>
      \end{lstlisting}
      \end{tcolorbox}
    
    \end{tcolorbox}
    }
  \caption{\textbf{Task example.} \emph{Picking up trash}: initial and
    final simulation frames (top); generated scene analysis and BT in XML (bottom).}
  \label{fig:end_to_end}
\end{figure}

Regarding proprietary baselines, GPT-5 saturates the benchmark, serving as a performance ceiling against these lightweight models. 
While Claude also produces syntactically valid trees in every trial, its success rate is substantially lower than GPT-5, reaching 40\% with CoT and 53\% in
zero-shot prompting. Interestingly, zero-shot prompting performs better for Claude
than the supplied workflow, suggesting that the additional decomposition can
occasionally constrain rather than improve its planning. Figure~\ref{fig:end_to_end} shows a representative easy task: from a single RGB
frame, Gemma-3 identifies all three target objects and generates a grounded plan
that satisfies all BDDL predicates on the first attempt.

The gap between Zero-Shot (ZS) and Chain-of-Thought (CoT) provides several insights into model capabilities. On easy tasks, the
two strategies produce comparable results due to the required plan being a direct
repetition of patterns already present in the training data. The difference emerges on medium and hard tasks, where the model must actively
manage container states or respect implicit ordering constraints that are not
stated in the goal. The CoT prompt externalises this reasoning into the structured
workflow, acting as scaffolding that benefits models capable of
generating valid XML but struggling with multi-step planning. Conversely, for SmolVLM2, the additional context tokens prove to be counterproductive, with CoT actually reducing BT validity.

\begin{table}[t]
  \centering
  \caption{Simulation results on the 15 tasks of BEHAVIOR-1K, either Zero-Shot (ZS) or Chain-of-Thought (CoT)}
  \label{tab:aggregate_results}
  \scriptsize
  \begin{tabular}{lllccc}
    \toprule
    \textbf{Model} & \textbf{Var.} & \textbf{Pr.}
      & \textbf{BT Valid} & \textbf{SR} & \textbf{Pass@3} \\
    \midrule
    GPT-5       & ---  & CoT & 100\% & 100\% & 100\% \\
                &      & ZS  & 100\% & 100\% & 100\% \\
    \midrule
    Claude Opus 4.8 & --- & CoT & 100\% & 40\% & 53\% \\
                    &     & ZS  & 100\% & 53\% & 67\% \\
    \midrule
    Gemma-3 4B  & FT   & CoT & 100\% & \textbf{87\%}  & \textbf{93\%} \\
                &      & ZS  & 100\% & \textbf{73\%}  & \textbf{80\%}  \\
                & Base & CoT & 0\%   & 0\%   & 0\%   \\
                &      & ZS  & 0\%   & 0\%   & 0\%   \\
    \midrule
    Qwen2.5-VL  & FT   & CoT & 100\% & 67\%  & 87\%  \\
                &      & ZS  & 100\% & 47\%  & 67\%  \\
                & Base & CoT & 0\%   & 0\%   & 0\%   \\
                &      & ZS  & 0\%   & 0\%   & 0\%   \\
    \midrule
    SmolVLM2    & FT   & CoT & 7\%   & 0\%   & 0\%   \\
                &      & ZS  & 27\%  & 0\%   & 7\%   \\
                & Base & CoT & 0\%   & 0\%   & 0\%   \\
                &      & ZS  & 0\%   & 0\%   & 0\%   \\
    \bottomrule
  \end{tabular}
\end{table}

\subsection{Ablation Studies}
\label{subsec:ablations}
In this section, we quantify the impact of three key components in our pipeline: dataset augmentation, the addition of vision and prompt guidance, using Gemma-3.

\noindent\textbf{Dataset augmentation.}
To evaluate the contribution of each augmentation step, we fine-tuned Gemma-3 4B on four different training sets: no augmentation, structural augmentation only, lexical augmentation only and both augmentations.
The results are reported in Table~\ref{tab:augmentation_ablation}.
We report the same offline metrics as in Table~\ref{tab:offline_results}, except for the inference time which is not affected by this ablation study. 
The results demonstrate that both augmentations are beneficial, with structural augmentation having a larger impact than lexical augmentation. 
When applied independently, structural augmentation outperforms lexical augmentation, improving linear structural compliance by 9\%, decorator compliance by 33\%, and action Jaccard by 4\%.
The cumulative effect of both augmentations is the optimal combination, with a perfect score on linear structural compliance and the highest scores on both decorator compliance and action Jaccard.

\begin{table}[t]
  \centering
  \caption{Ablation on dataset augmentation using Gemma-3 on the held-out test split
  ($N=228$).}
  \label{tab:augmentation_ablation}
  \scriptsize
  \setlength{\tabcolsep}{3.5pt}
  \begin{tabular}{lccccc}
    \toprule
    & \multicolumn{2}{c}{\textbf{Validity (\%)}}
      & \multicolumn{2}{c}{\textbf{Structural compliance (\%)}} & \\
    \cmidrule(lr){2-3} \cmidrule(lr){4-5}
    \textbf{Training aug.} & \textbf{XML} & \textbf{BT-CPP}
      & \textbf{Linear} & \textbf{Decorator} & \textbf{Action Jaccard} \\
    \midrule
    None       & \textit{65} & \textit{65} & \textit{109 / 152} & \textit{38 / 76} & \textit{$0.898{\pm}0.079$} \\
    Structural & \textit{93} & \textit{91} & \textit{135 / 152} & \textit{67 / 76} & \textit{$0.958{\pm}0.088$} \\
    Lexical    & \textit{73} & \textit{73} & \textit{122 / 152} & \textit{42 / 76} & \textit{$0.922{\pm}0.088$} \\
    Both       & \textbf{100} & \textbf{100} & \textbf{152 / 152} & \textbf{69 / 76} & $\mathbf{0.971}{\pm}0.129$ \\
    \bottomrule
  \end{tabular}
\end{table}

\noindent\textbf{Visual modality.}
We also isolate the contribution of the visual modality by fine-tuning a version of Gemma-3 4B on the same dataset but without the visual input, so that the model is conditioned only on the task instruction and the list of allowed actions.
We evaluated it on the same fifteen BEHAVIOR-1K tasks, under both prompting strategies and with the same metrics as in Table~\ref{tab:aggregate_results} except for Pass@3.
The results are reported in Table~\ref{tab:modality_ablation}.
Both variants generate a valid BT, suggesting that the input modality does not affect the syntactic correctness of the BT and that text is enough to generate valid trees.
However, visual input is crucial for task success, as the model must ground its plan in the actual objects present in the scene.
Removing it halves the success rate, from 87\% to 40\% with CoT and from 73\% to 33\% in zero-shot.
This demonstrates the impact of adding visual input, as the text-only model produces a plan that is consistent with the task instruction but not grounded with the objects actually present in the scene.

\begin{table}[t]
  \centering
  \caption{Modality ablation on BEHAVIOR-1K.}
  \label{tab:modality_ablation}
  \scriptsize
  \begin{tabular}{llccc}
    \toprule
    \textbf{Model} & \textbf{Input} & \textbf{Pr.} & \textbf{BT Validity} & \textbf{SR} \\
    \midrule
    Gemma-3 4B (FT) & Text          & CoT & 100\% & 40\% \\
                    &               & ZS  & 100\% & 33\% \\
    \midrule
    Gemma-3 4B (FT) & Text + vision & CoT & 100\% & \textbf{87\%} \\
                    &               & ZS  & 100\% & 73\% \\
    \bottomrule
  \end{tabular}
\end{table}

\begin{figure}[t]
  \centering
  \includegraphics[width=\columnwidth]{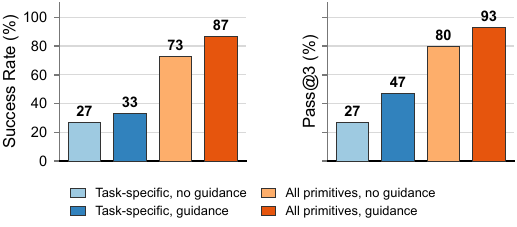}
  \caption{Ablation on action vocabulary and prompt guidance on BEHAVIOR-1K. BT validity is unaffected and remains 100\% in all four configurations.}
  \label{fig:aux_ablation}
\end{figure}

\noindent\textbf{Action vocabulary and prompt guidance.}
Finally, in Figure~\ref{fig:aux_ablation} we examine the impact of the action vocabulary and prompt guidance.
The action vocabulary can either be a subset of primitives specific to the task or the full set of primitives.
Subsequently, the prompt guidance can either be provided as the natural language \texttt{Workflow} of the CoT prompt, which lists the required primitives and their preconditions, or not provided at all, as in the Zero-Shot prompt.
BT validity achieves a perfect score in all four configurations, confirming that neither factor affects the syntactic quality of the tree, but is instead a result of the augmentation in Table~\ref{tab:augmentation_ablation}.
The full action vocabulary is the dominant factor, improving SR and Pass@3 by 46\% and 53\% without guidance, and by 54\% and 46\% with guidance, respectively.
Overall, prompt guidance yields smaller gains and improves the success rate and Pass@3 only marginally when compared with the task-specific subset.
These results suggest that the model mainly benefits from retaining all primitives, as pruning the vocabulary can actually omit required primitives and constrain the model to incomplete plans, while prompt guidance provides an additional improvement without being the primary driver of task success.

\subsection{Failure Analysis}
\label{subsec:failure_analysis}

Although the exact type of error varies by model, first-attempt failures mainly concentrate on the following tasks: \emph{moving boxes to storage}, \emph{preparing lunch box}, \emph{putting away Halloween decorations}, and \emph{carrying in groceries}.
\emph{Carrying in groceries} is the most challenging task, with GPT-5 being the only model to solve it under both ZS and CoT. 
For this reason, we investigate a failure case for this task in Figure~\ref{fig:model_comparison} by comparing the CoT outputs of all five models.
For instance, this hard task requires a complex sequence: placing a sack beside the refrigerator, opening the fridge with empty hands, storing a tomato and a carton of milk inside, and finally closing both the fridge and the car. GPT-5 understands the implicit physical constraint: it places the sack, frees its hands, and only then opens the fridge.
Claude produces a valid plan but uses the unsupported \texttt{destination} parameter instead of the \texttt{obj} for placement nodes, causing execution to abort at the first \texttt{PLACE\_NEXT\_TO} action.
Gemma-3 produces valid XML containing the correct objects and actions, but violates the sequential ordering: it
grasps the tomato before opening the fridge, which fails because the
gripper is already occupied. Qwen similarly generates valid XML, but the plan is
semantically ambiguous: it attempts to open the fridge while still holding the
sack, uses an incorrect placement action, and attempts to grasp the car.
Conversely, SmolVLM2 never reaches the planning stage because it fails to produce a
well-formed XML. Ultimately, each model fails at a different
level. As the number of parameters decreases from GPT-5 to SmolVLM2, the errors shift systematically from logical planning flaws to basic syntax failures, reflecting the relationship between model scale and the type of failure.

\begin{figure}[ht]
\centering
\begin{tcolorbox}[enhanced, width=\columnwidth,
  boxrule=0.5pt, colframe=bluePoli!40, colback=bluePoli!3,
  arc=2pt, outer arc=2pt,
  top=0pt, bottom=0pt, left=0pt, right=0pt]

\begin{tcolorbox}[enhanced,
  colback=bluePoli!5, colframe=bluePoli!5,
  fonttitle=\sffamily\bfseries\scriptsize\color{black},
  title={\textbf{GPT-5}~{\normalsize\textbf{\textcolor{green!60!black}{\checkmark}}}},
  boxrule=0pt, arc=0pt,
  top=1pt, bottom=1pt, left=4pt, right=4pt,
  toptitle=1pt, bottomtitle=1pt,
  before skip=0pt, after skip=0pt]
{\fontsize{5.5}{7}\selectfont
\lstinline[style=btxml,basicstyle=\fontsize{5.5}{7}\ttfamily\color{bodycol}]|<Action ID="PLACE_NEXT_TO" obj="electric_refrigerator"/>|\\
\lstinline[style=btxml,basicstyle=\fontsize{5.5}{7}\ttfamily\color{bodycol}]|<Action ID="OPEN" obj="electric_refrigerator"/>|
\;{\scriptsize\textit{$\leftarrow$ hands free, succeeds}}\\
\lstinline[style=btxml,basicstyle=\fontsize{5.5}{7}\ttfamily\color{bodycol}]|<Action ID="GRASP" obj="beefsteak_tomato"/>|\\[-1pt]
\hspace*{1.2em}{\ttfamily\color{bodycol}\fontsize{5.5}{7}\selectfont \dots}\\[-1pt]
\lstinline[style=btxml,basicstyle=\fontsize{5.5}{7}\ttfamily\color{bodycol}]|<Action ID="CLOSE" obj="car"/>|
}
\end{tcolorbox}

\vspace{1pt}\textcolor{bluePoli!40}{\hrule}\vspace{1pt}

\begin{tcolorbox}[enhanced,
  colback=bluePoli!3, colframe=bluePoli!3,
  fonttitle=\sffamily\bfseries\scriptsize\color{black},
  title={\textbf{Claude Opus 4.8}~{\normalsize\textcolor{red!70!black}{$\times$}}~{\normalfont\scriptsize parameter error}},
  boxrule=0pt, arc=0pt,
  top=1pt, bottom=1pt, left=4pt, right=4pt,
  toptitle=1pt, bottomtitle=1pt,
  before skip=0pt, after skip=0pt]
{\fontsize{5.5}{7}\selectfont
\lstinline[style=btxml,basicstyle=\fontsize{5.5}{7}\ttfamily\color{bodycol}]|<PLACE_NEXT_TO destination="electric_refrigerator"/>|
\;{\scriptsize\textit{$\leftarrow$ \texttt{obj} required}}
}
\end{tcolorbox}

\vspace{1pt}\textcolor{bluePoli!40}{\hrule}\vspace{1pt}

\begin{tcolorbox}[enhanced,
  colback=bluePoli!3, colframe=bluePoli!3,
  fonttitle=\sffamily\bfseries\scriptsize\color{black},
  title={\textbf{Gemma-3 4B}~{\normalsize\textcolor{red!70!black}{$\times$}}~{\normalfont\scriptsize ordering error}},
  boxrule=0pt, arc=0pt,
  top=1pt, bottom=1pt, left=4pt, right=4pt,
  toptitle=1pt, bottomtitle=1pt,
  before skip=0pt, after skip=0pt]
{\fontsize{5.5}{7}\selectfont
\lstinline[style=btxml,basicstyle=\fontsize{5.5}{7}\ttfamily\color{bodycol}]|<Action ID="GRASP" obj="beefsteak_tomato"/>|
\;{\scriptsize\textit{$\leftarrow$ before opening}}\\
\lstinline[style=btxml,basicstyle=\fontsize{5.5}{7}\ttfamily\color{bodycol}]|<Action ID="OPEN" obj="electric_refrigerator"/>|
\;{\scriptsize\textit{$\leftarrow$ fails: hands full}}
}
\end{tcolorbox}

\vspace{1pt}\textcolor{bluePoli!40}{\hrule}\vspace{1pt}

\begin{tcolorbox}[enhanced,
  colback=bluePoli!3, colframe=bluePoli!3,
  fonttitle=\sffamily\bfseries\scriptsize\color{black},
  title={\textbf{Qwen2.5-VL-3B}~{\normalsize\textcolor{red!70!black}{$\times$}}~{\normalfont\scriptsize ordering and grounding errors}},
  boxrule=0pt, arc=0pt,
  top=1pt, bottom=1pt, left=4pt, right=4pt,
  toptitle=1pt, bottomtitle=1pt,
  before skip=0pt, after skip=0pt]
{\fontsize{5.5}{7}\selectfont
\lstinline[style=btxml,basicstyle=\fontsize{5.5}{7}\ttfamily\color{bodycol}]|<Action ID="OPEN" obj="electric_refrigerator"/>|
\;{\scriptsize\textit{$\leftarrow$ hands full}}\\
\lstinline[style=btxml,basicstyle=\fontsize{5.5}{7}\ttfamily\color{bodycol}]|<Action ID="GRASP" obj="car"/>|
\;{\scriptsize\textit{$\leftarrow$ wrong target object}}
}
\end{tcolorbox}

\vspace{1pt}\textcolor{bluePoli!40}{\hrule}\vspace{1pt}

\begin{tcolorbox}[enhanced,
  colback=bluePoli!3, colframe=bluePoli!3,
  fonttitle=\sffamily\bfseries\scriptsize\color{black},
  title={\textbf{SmolVLM2-500M}~{\normalsize\textcolor{red!70!black}{$\times$}}~{\normalfont\scriptsize invalid XML}},
  boxrule=0pt, arc=0pt,
  top=1pt, bottom=1pt, left=4pt, right=4pt,
  toptitle=1pt, bottomtitle=1pt,
  before skip=0pt, after skip=0pt]
{\fontsize{5.5}{7}\selectfont
\lstinline[style=btxml,basicstyle=\fontsize{5.5}{7}\ttfamily\color{bodycol}]|</Sequence>|
\;{\scriptsize\textit{$\leftarrow$ tree closes early}}\\
\lstinline[style=btxml,basicstyle=\fontsize{5.5}{7}\ttfamily\color{bodycol}]|<Action ID="NAVIGATE_TO" .../>|
\;{\scriptsize\textit{$\leftarrow$ outside tree}}
}
\end{tcolorbox}

\end{tcolorbox}
\caption{Failure comparison on the \emph{carrying in groceries} task with CoT.}
\label{fig:model_comparison}
\end{figure}

Across the entire evaluation phase, two failure patterns emerge. The first is
the systematic violation of physical preconditions, such as attempting to place an object inside a closed container (e.g., \texttt{PLACE\_INSIDE} before \texttt{OPEN}) or attempting to open a door while the gripper is occupied (e.g., \texttt{GRASP} before
\texttt{OPEN}). The second is object hallucination, where models
generate new target names or confuse object suffixes, even when exact lists are explicitly provided in the prompt. Crucially, both patterns become more frequent as model size decreases.

\section{Conclusions}
\label{sec:conclusions}

This work shows that compact vision-language models, trained with structured supervision, can generate grounded and executable behavior trees without relying on large proprietary models. 
By constructing a multimodal BT dataset from Open X-Embodiment and fine-tuning open-source VLMs with QLoRA, we obtain models that produce ROS2-compatible BTs from a single RGB observation and a natural-language instruction. 
On 15 BEHAVIOR-1K tasks, Gemma-3 with CoT prompting achieves 87\% success and 93\% Pass@3, outperforming Claude Opus 4.8 and approaching the performance of GPT-5, with only 4B parameters. 
The ablations show that vision primarily improves semantic task success, whereas dataset augmentation improves structural reliability. 
The remaining failures are concentrated in longer-horizon tasks involving implicit physical preconditions and object grounding. 
Future work will focus on keeping the visual channel active during execution and enriching the training set with longer-horizon demonstrations.

\bibliographystyle{IEEEtran}
\bibliography{bibliography}

\addtolength{\textheight}{-12cm} 

\end{document}